# Visual Reasoning at Urban Intersections: Fine-Tuning GPT-4o for Traffic Conflict Detection

Sari Masri
*Department of Natural, Engineering and Technology Sciences, Faculty of Graduate Studies*
*Arab American University*
Ramallah, Palestine
s.masri3@student.aaup.edu

Huthaifa I. Ashqar
*AI and Data Science Department*
*Arab American University*
Ramallah, Palestine
huthaifa.ashqar@aaup.edu

Mohammed Elhenawy
*CARRS-Q*
*Centre for Data Science*
*Queensland University of Technology*
Brisbane, Australia
mohammed.elhenawy@qut.edu.au

***Abstract*—Traffic control in unsignalized urban intersections presents significant challenges due to the complexity, frequent conflicts, and blind spots. This study explores the capability of leveraging Multimodal Large Language Models (MLLMs), such as GPT-4o, to provide logical and visual reasoning by directly using birds-eye-view videos of four-legged intersections. In this proposed method, GPT-4o acts as intelligent system to detect conflicts and provide explanations and recommendations for the drivers. The fine-tuned model achieved an accuracy of 77.14%, while the manual evaluation of the true predicted values of the fine-tuned GPT-4o showed significant achievements of 89.9% accuracy for model-generated explanations and 92.3% for the recommended next actions. These results highlight the feasibility of using MLLMs for real-time traffic management using videos as inputs, offering scalable and actionable insights into intersections traffic management and operation. Code used in this study is available at https://github.com/sarimasri3/Traffic-Intersection-Conflict-Detection-using-images.git.**

## I. INTRODUCTION

Urban intersections are highly challenging due to their unpredictability and dynamism, especially in cases of unsignalized intersections. Interactions often occur among motor vehicles and other road users in such areas. Managing these intersections is difficult because of the high frequency of accidents, blind spots, and the lack of a defined control mechanism that characterizes them [1], [2], [3]. Traditional traffic management systems are static mechanisms that address problems only as they arise. However, technological advancements now demand new adaptive systems capable of detecting potential conflicts and reacting to them in real time [4], [5]. There is a pressing challenge to develop inventive solutions that detect conflicts and provide immediate plans for resolution.

Recently, the field has witnessed advancements in artificial intelligence, particularly in the context of Multimodal Large Language Models (MLLMs), such as GPT-4o. The advantage of MLLMs lies in their ability to perform logical reasoning, contextual understanding, and decision-making [6], [7]. When integrated with video data analysis, these capabilities offer the potential to revolutionize traffic management by enabling intelligent traffic control [8], [9]. In this context, frame analysis, feeding the first three consecutive frames of video data in the correct order to the model, proves effective for conflict detection. The movement in the video clearly indicates which frame is the first, second, and third, ensuring the model processes them in the intended sequence. By analyzing sequentially extracted patterns and interactions captured by drones, Frame Analysis systematically identifies potential conflicts and categorizes traffic conditions as either conflict or non-conflict.

This paper investigates the use of MLLM-based traffic control for unsignalized intersections such as those provided by GPT-4o. The system architecture comprises drone-captured video footage for detecting and classifying conflicts, making detailed explanations and recommendation actions for drivers. This study also includes iterative-prompt optimization and fine-tuning to enhance conflict detection and responses. This strategy aims to leverage the intelligent part of the language models to be more adaptive towards complicated intersection scenarios to provide realistic and actionable solutions dynamically [10], [11].

## II. LITERATURE REVIEW

MLLMs have emerged as powerful tools in managing traffic and self-driving vehicles, providing responsive, adaptable, and comprehensible solutions [12], [13], [14]. They allow it possible to draw specific suggestions, which appear useful across varied user categories such as drivers, engineers, and policy planners. For instance, they have facilitated the development of internet-enabled traffic lights along intelligent routes [15], [16], [17]. Research articles examining machine learning techniques in transportation systems have outlined their advantages and disadvantages because they chart out the future focus areas [18], [19].

Recent studies have classified the use of Large Language Models (LLMs) in self-driving systems into four primary categories: planning, perception, question-answering, and generation. The study LLM4DRIVE in [19] highlights obstacles related to these applications' clarity, scalability, and practicability. More importantly, it argues strongly for developing reliable datasets and adequately interpretable models that could enhance trustworthiness within these systems. In [20] similarly discussed how autonomous driving technologies have evolved from sensor-based approaches to sophisticated deep learning methods using examples of vision foundation models (VFMs) designed for better perception planning decision-making but lack application in complex, unsignalized intersections, which present unique challenges in conflict resolution and real-time decision-making [21].

New designs show how else LLMs could predict future traffic conditions as well as control driverless vehicles. In their paper, [22] introduced several LLM-based frameworks with the sequence as well as graph embedding layers, which resulted in a good performance in few-shot learning tasks on historical data analysis DriveMLM was a framework developed in [17] to align multimodal LLMs with behavioral planning states thereby facilitating incorporation of language-based decisions with vehicle control commands in simulators, while in [20] was about making human-vehicle interactions more intuitive through LLMs processing natural language commands. Additionally, [17] introduced AccidentGPT to enable reconstruction of traffic accidents. However, most existing works focus on structured environments with predefined datasets and lack real-time adaptability for intersection-level control. Moreover, this sensory data integration with LLMs has significantly improved the ability of autonomous systems to perceive information. For example, in [23] combined LLMs with LiDAR and radar information to enhance object distinction and following, while in [24] predicted human movement patterns using LLMs that analyze context clues and visual information. Similarly, [25] explored driver-vehicle interaction via motion and voice command interpretation, and [26] used real-time car dashboard videos to identify dangers such as abrupt lane changes and barriers, thus improving general road safety. Despite these advances, current studies do not fully integrate real-time traffic perception with decision-making strategies for unsignalized intersections.

Explainability has also become a focal point in deploying LLMs for critical systems such as autonomous driving. Integrating these technologies promotes confidence in road user behavior predictions using knowledge graphs and retrieval-augmented generation (RAG) under LLMs [27] Multimodal LLMs, in combination with holistic traffic foundation models, can be used for better transportation data analysis [28], while reinforcement learning helps address hard-driving scenarios like uncontrolled crossroads [29].

This study addresses these gaps by being one of the first to apply fine-tuned MLLMs like GPT-4o to control traffic at unsignalized intersections using birds-eye-view videos. The proposed approach not only recognizes conflicts but also classifies them and provides detailed, explainable justifications with actionable suggestions for vehicles. This dynamic and adaptive decision-making framework enhances intersection safety and efficiency.

## III. Materials and Methods

The framework illustrated in Fig. 1 is designed for detecting conflicts from videos extracted from drones, utilizing preprocessing steps, and feeding the data to a MLLM, specifically GPT-4o. The process begins with frame extraction, where the system collects and resizes three frames from the drone's camera every 0.5 seconds. These frames are then subjected to the vehicle detection phase, where the model evaluates the scene for potential conflicts.

### A. Data Collection and Labeling

The dataset used in this study has been collected from publicly available drone footage for unsignalized urban intersections under different traffic conditions captured with regard to other vehicle types, such as cars, buses, vans, trucks, bikes, and motorcycles [30]. Video data has been segmented into frames, extracting three sequential frames for each scenario at 0.5 seconds intervals. Each observation included these three frames with a conflict label stating whether or not a potential conflict existed between observed vehicles. Labels were annotated manually based on observed movements, priority rule violations, and likelihood of conflict according to the overlap of vehicle trajectories. The final dataset had 700 labeled observations equally distributed between conflict and no conflict. These observations were labeled manually by observing the last frame of each scenario and logically determining whether a conflict existed in the intersection. The observations were divided into three subsets used for training and evaluating the model: 504 observations for the training set (252 conflicts, 252 no-conflict); 56 observations for the validation set (28 conflicts, 28 no-conflict); and 140 observations for the test set (70 conflicts, 70 no-conflict). Such a balanced division has lent itself to reliable training, fine-tuning, and evaluation of the models used in this study.

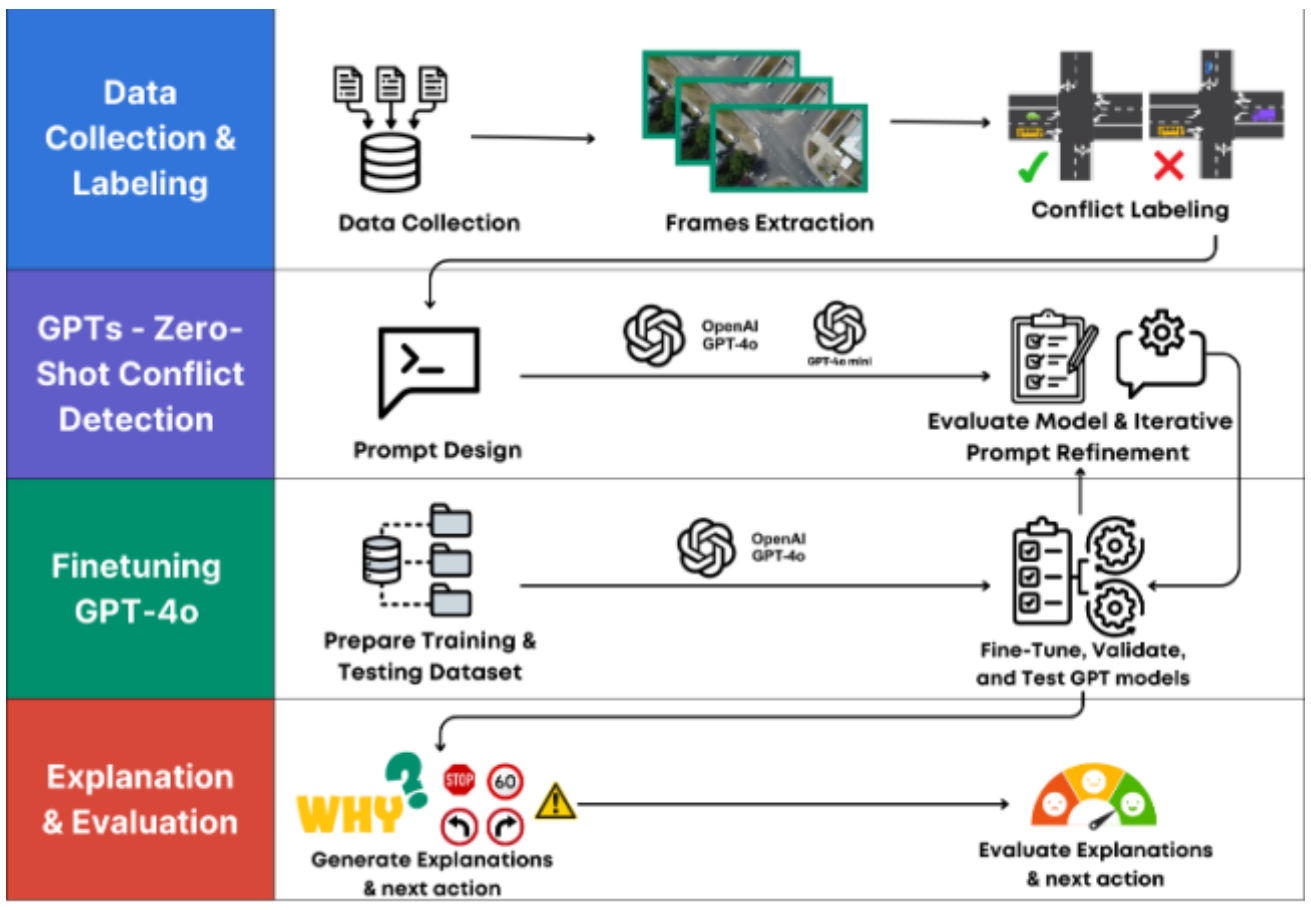

Fig. 1. Flowchart of the proposed framework.

### B. Prompt Design

Two structured prompts were designed to help LLMs develop for traffic conflict detection—the first prompt concerns real-time traffic situations, which would entail interpreting vehicle interactions at an urban intersection. The second prompt elaborated more about the layout for the intersection, detailing the number of lanes along which directions, left turns, right turns, or straight-ahead movements were to be taken. It also included information on frames and intervals on a temporal aspect of traffic dynamics. These structured prompts allow the LLM to analyze the flow of events and possible conflicts that transpire within them. Prompt 1 (P1) and prompt 2 (P2) are shown in TABLE I.

TABLE I. Prompt Structure.

| |
|---|
| *Prompt 1 (P1)*<br>You are a traffic control AI analyzing drone footage of a four-way intersection with two main roads and two sub-roads. Analyze frames showing moving vehicles before the intersection to detect potential conflicts.<br>- Answer strictly only with "yes" or "no" in lowercase. |
| *Prompt 2 (P2)*<br>Analyze three sequential overhead images of a four-leg intersection, 0.5s apart. West-East (main) road has priority. Two lanes each way on main road, with dedicated turn lanes. North-South (sub) road has single lanes with shared turn/crossing. Ignore parked cars. Focus on moving vehicles intending to cross the intersection or turn. If, after all three frames, any unresolved conflict may occur (e.g., priority vehicle and another vehicle potentially entering the same space), answer 'yes' (lowercase); otherwise, answer 'no' (lowercase).<br>- Answer strictly only with "yes" or "no" in lowercase to detect conflicts. |

### C. Zero-Shot (ZS) Evaluation

Zero-shot performance testing of the GPT-4o and GPT-4o-mini pre-trained models mainly aimed at critical global performance assessment in detection of traffic conflict. Zero-shot model was used as a baseline. This was done using the two prompts on 140 test observations (70 conflict, 70 no-conflict). Performance metrics were used including accuracy, precision, recall, and F1 score. This evaluation will serve as the groundwork for testing the extent of improvement by incorporating prompt designing and personalized models.

### D. Fine-Tuning Process

Fine-tuned GPT-4o enhanced traffic incident identification abilities based on structured prompts. A training set containing 504 evenly divided observations in conflict and no-conflict situations was used to set the optimal parameters for the model. Another 56 new observations were used as validation to check on training performance and fine-tune hyperparameters to avoid overfitting. The remining 140 observations were used for testing. The performance scores included accuracy, precision, recall, and F1-score metrics, reflecting the model's effectiveness. It was found that fine-tuning brought significant improvements over the zero-shot baseline, suggesting that specific training for the task and carefully designed prompts might enhance the ability of conflict detection within the model.

### E. Model Evaluation Metrics

Model performance evaluation relied on four leading indicators: accuracy, precision, recall, and F1 score, which are usually used for evaluating classification tasks. They were picked to provide a complete assessment of the ability of the model to identify conflicts and avoid misclassification.

### F. Manual Evaluation of Explanations and Recommendations

Explanations and suggestions for handling the conflicts identified within the GPT-4o model while fine-tuning were extended during the fine-tuning process. For example, if a conflict was reported, the model suggested changing traffic signal timings or vehicle rerouting to avoid accidents. These explanations and recommendations were then manually evaluated by a panel of three traffic management experts to ensure that the message was understood correctly, aligned with real-world traffic control practices, and met safety and clarity standards.

In contrast to external approaches, the explanations and suggestions were native, rather than external guidelines for interpretation, made by a fine-tuned GPT-4o model during detection and detection processes. This enabled a smooth delivery of both detections and actionable insights using the system workflow.

The review was carried out manually, and the model's explanatory quality and recommendation features were assessed by these three experts, each with experience in traffic planning and control. They scored the output on a 0–10 scale based on three key parameters: clarity (whether the explanation was understandable), accuracy (how well the suggestion aligned with traffic rules and best practices), and practical relevance (the feasibility of implementing the recommendation in a real-world scenario). The final scores were averaged to quantify the overall interpretability and effectiveness of the model's outputs. This kind of study enabled both quantitative and qualitative insights into how interpretable the system was and how it might make decisions.

## IV. Experimental Results

### A. Model Performance Metrics

The accuracy, precision, recall, and F1-score results are shown in TABLE II and Fig. 2. The fine-tuned GPT-4o model performed best when using Prompt 2 (P2), achieving an accuracy of 77.14%. In comparison, the same model reached 67.14% accuracy with Prompt 1 (P1). This improved performance with P2 is attributed to the prompt's ability to provide more accurate and detailed information, helping the model better understand traffic dynamics and predict future conflicts. To ensure a fair comparison, the few-shot model was retested on the same test dataset used for fine-tuning, which consisted of 140 samples.

In a zero-shot setting, GPT-4o achieved lower accuracies of 58.43% with P2 and 55.43% with P1. The smaller GPT-4o-mini model, which has limited capacity, performed even worse, with accuracies of 53.71% (P1) and 50.29% (P2). These results highlight the importance of fine-tuning and model size for handling complex traffic scenarios.

Beyond accuracy, the fine-tuned GPT-4o model with P2 showed strong overall performance: 78% precision, 77.5% recall, and an F1-score of 77%. With P1, these metrics dropped to 74.5% precision, 67% recall, and an F1-score of 64.5%. In zero-shot settings, GPT-4o struggled further, achieving only 61.5% precision, 58.5% recall, and a 55.5% F1-score. The GPT-4o-mini model's F1-score was particularly low, at just 35% with P2, reflecting its inability to handle the complexity of traffic data.

These results confirm the critical role of fine-tuning, thoughtful prompt design, and model size in improving performance. Among all tests, Prompt 2 consistently delivered the best results, outperforming both Prompt 1 and the zero-shot setups across all metrics. This demonstrates that well-designed prompts and appropriately fine-tuned models are essential for solving complex traffic management systems.

TABLE II. Performance Metrics of Zero-Shot and Fine-Tuned Models for Traffic Conflict Detection Using Different Prompts

| *Model* | *Accuracy* | *Precision* | *Recall* | *F1-Score* |
|---|---|---|---|---|
| GPT-4o fine-tuned P2 | 0.771 | 0.771 | 0.771 | 0.771 |
| GPT-4o fine-tuned P1 | 0.78 | 0.78 | 0.78 | 0.78 |
| GPT-4o ZS P2 | 0.77 | 0.77 | 0.77 | 0.77 |
| GPT-4o ZS P1 | 0.77 | 0.77 | 0.77 | 0.77 |
| GPT-4o-mini ZS P1 | 0.671 | 0.671 | 0.671 | 0.671 |
| GPT-4o-mini ZS P2 | 0.74 | 0.74 | 0.74 | 0.74 |

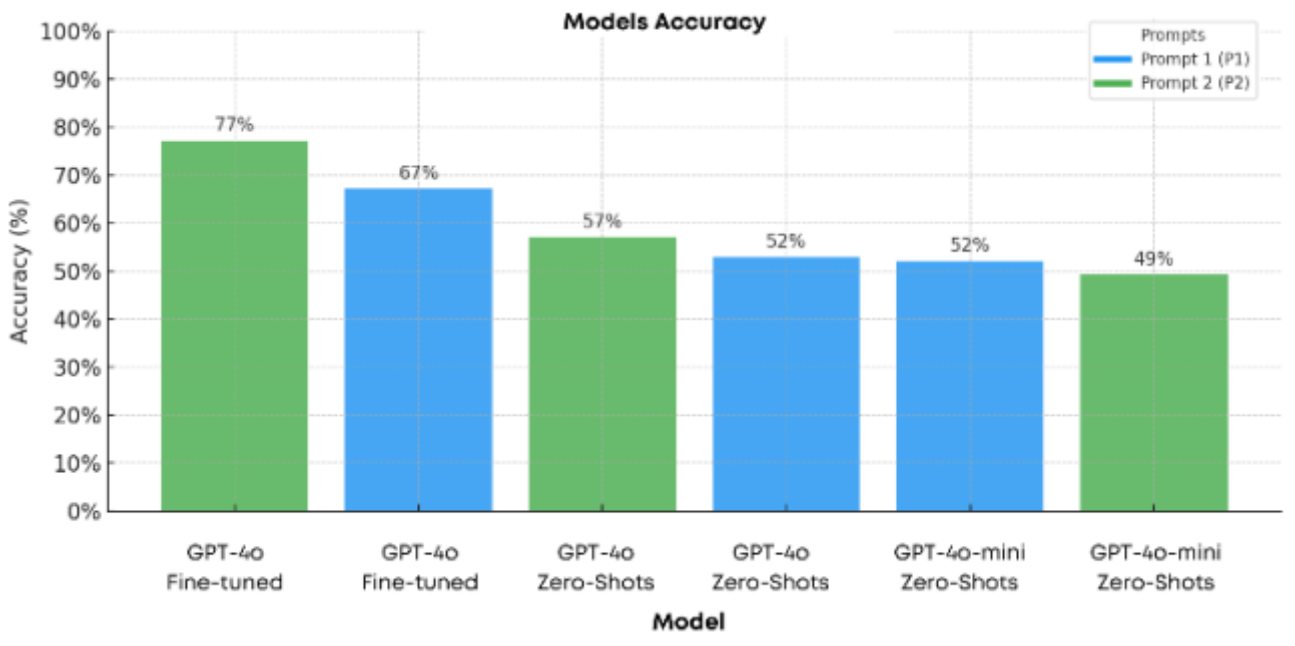

Fig. 2. Accuracy Comparison of Zero-Shot and Fine-Tuned Models Using Prompt 1 and Prompt 2.

### B. Confusion Matrices

The confusion matrices in Fig. 3 illustrate how the GPT-4o model performed when fine-tuned with two different prompts. For Prompt 1, the model recorded 66 TN and 28 TP, but it also produced 42 FN and 4 FP. While the model was fairly good at identifying "no conflict" situations, the high number of false negatives suggests it struggled to detect certain conflict scenarios due to the lack of precise context in Prompt 1.

In contrast, Prompt 2 significantly improved the model's ability to identify conflicts. With this prompt, the model achieved 60 TN and 48 TP, while false negatives dropped to 22, and false positives increased slightly to 10. This performance highlights Prompt 2’s ability to help the model better interpret vehicle interactions and prioritize traffic rules.

These results emphasize how well-designed prompts can enhance the model's effectiveness. By including richer contextual cues, such as vehicle priorities and movement dynamics, Prompt 2 enabled the model to make more accurate predictions and handle traffic scenarios more effectively.

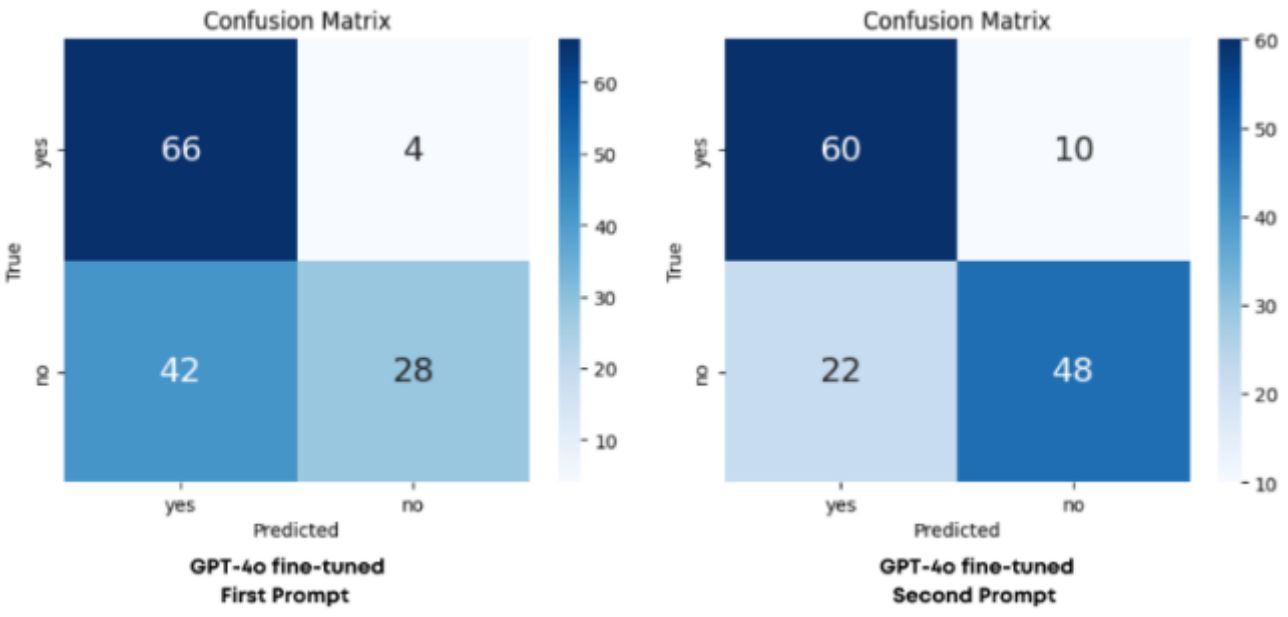

Fig. 3. Confusion Matrices for GPT-4o Fine-Tuned Model Using First and Second Prompts for Traffic Conflict Detection.

## C. Manual Evaluation

The quality of the model’s explanations was manually evaluated, focusing on clarity and usefulness. On a 10-point scale, the explanations scored an average of 8.99, while the recommendations scored slightly higher at 9.23. This reflects the practical value of the recommendations in addressing issues like conflict resolution and ensuring safe traffic flow. These results highlight the model’s potential for real-world traffic management applications, as illustrated in Fig. 4. Examples of the fine-tuned GPT-4o model’s outputs are shown in Fig. 5, further demonstrating its effectiveness.

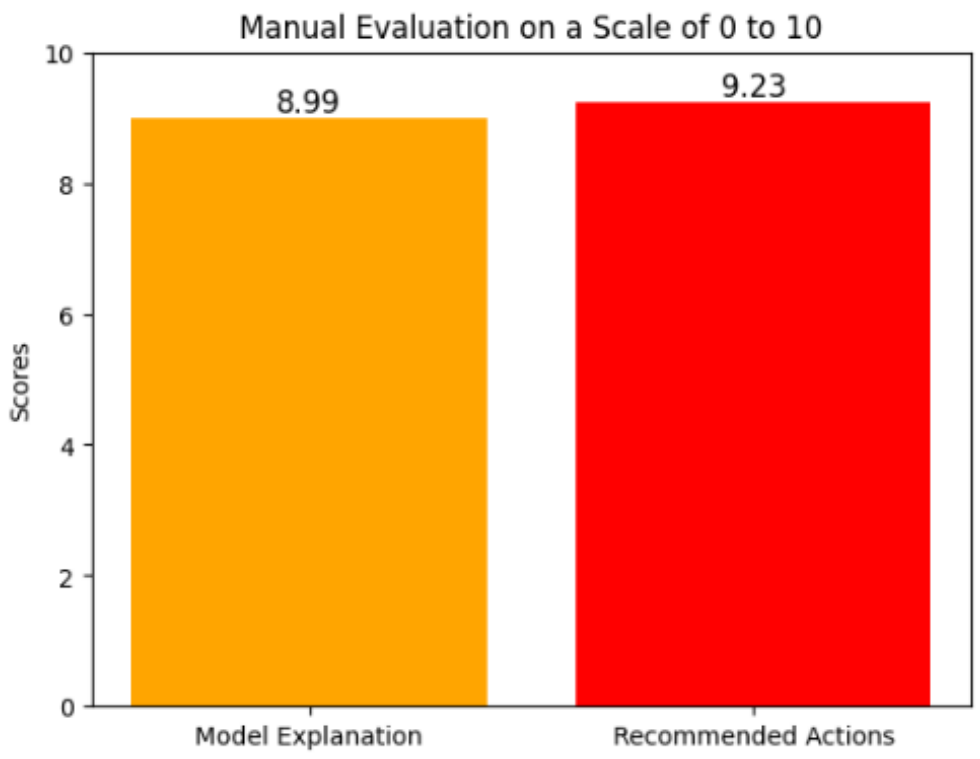

Fig. 4. Manual Evaluation Scores for Model Explanations and Recommended Actions on a Scale of 0 to 10.

Fig. 5 showcases examples of the fine-tuned GPT-4o model's outputs in analyzing traffic scenarios. Each row provides a clear demonstration of the model’s ability to detect conflicts, generate detailed explanations, and recommend actions to improve traffic safety.

The "Conflict detected" cases highlight the model's capability to identify potential issues, such as vehicles blocking intersections or unsafe vehicle maneuvers. The explanations include contextual insights, such as vehicle positions and priorities, while the recommendations focus on practical solutions, like adjusting traffic flow or assigning vehicle priorities.

For "No conflict detected" scenarios, the model accurately recognizes smooth traffic flow, providing logical justifications and confirming the absence of risks. The output demonstrates the model's usefulness in real-world applications, offering actionable insights for traffic management and ensuring safer and more efficient intersections. This underscores the fine-tuned GPT-4o model's strong performance in both conflict detection and resolution.

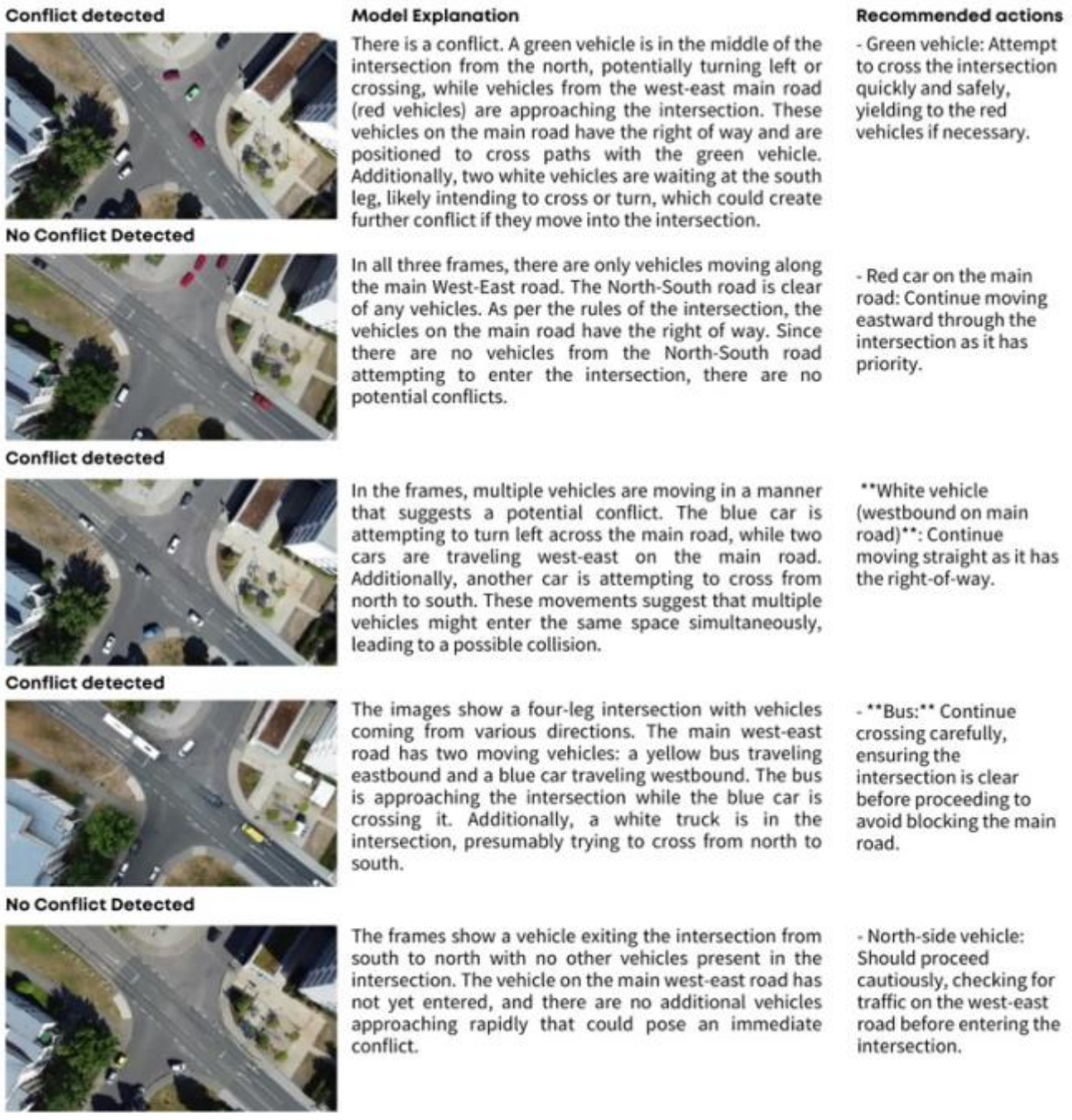

Fig. 5. Examples of the output of the fine-tuned GPT-4o model.

# V. Conclusion and Future Work

This study highlights the capacity of fine-tuned MLLMs, particularly GPT-4o, for traffic management at urban intersections. Through fine-tuning and effective prompt design, the model achieved about 77% accuracy, outperforming the zero-shot accuracy. Additionally, the model provided detailed explanations and actionable recommendations for traffic scenarios, with high scores for explanation quality (8.99) and recommendations (9.23) on a 10-point scale, underlining their practical applicability in addressing real-world traffic issues.

Comparisons with zero-shot performance (57.1% and 52.9% accuracy for the second and first prompts, respectively) and the smaller GPT-4o-mini model (accuracy of 52.1% and 49.3% for the first and second prompts, respectively) demonstrated the critical role of fine-tuning and prompt design in enhancing model effectiveness. Furthermore, the second prompt consistently outperformed the first prompt across all metrics, including precision (78% vs. 74.5%), recall (77.5% vs. 67%), and F1-score (77% vs. 64.5%). This emphasizes the importance of detailed contextual cues in prompt design, such as vehicle priorities and movement dynamics, for accurate predictions.

These results imply the potential of fine-tuned MLLMs like GPT-4o in real-time traffic management systems. The study highlights the importance of model size, fine-tuning, and prompt design in achieving practically high performance, laying a solid foundation for future advancements in using language models for traffic optimization and safety at unsignalized intersections.

Future work will continue by extending the dataset with observations from various traffic situations and improving data quality through richer contextualization. A promising avenue for further enhancing conflict detection is the joint use of MLLMs such as Gemini or LLaVA, which combine visual and textual information. Real-time deployment via live traffic feeds will also explore new MLLM architectures to further assess their validity. These steps aim to develop scalable, accurate, and context-aware smart solutions for traffic management.

## REFERENCES

[1] K. Bucsuházy, E. Matuchová, R. Zůvala, P. Moravcová, M. Kostíková, and R. Mikulec, "Human factors contributing to the road traffic accident occurrence," *Transportation Research Procedia*, vol. 45, pp. 555–561, 2020, doi: 10.1016/j.trpro.2020.03.057.

[2] H. I. Ashqar, Q. H. Q. Shaheen, S. A. Ashur, and H. A. Rakha, "Impact of risk factors on work zone crashes using logistic models and Random Forest," in *2021 IEEE International Intelligent Transportation Systems Conference (ITSC)*, IEEE, Sep. 2021, pp. 1815–1820. doi: 10.1109/ITSC48978.2021.9564405.

[3] I. Albool *et al.*, "Fuel consumption at signalized intersections: Investigating the impact of different signal indication settings," *Case Stud Transp Policy*, vol. 13, p. 101022, Sep. 2023, doi: 10.1016/j.cstp.2023.101022.

[4] M. Elhenawy, H. A. Rakha, and H. I. Ashqar, "Joint Impact of Rain and Incidents on Traffic Stream Speeds," *J Adv Transp*, vol. 2021, pp. 1–12, Jan. 2021, doi: 10.1155/2021/8812740.

[5] H. Rakha, A. Amer, and I. El-Shawarby, "Modeling Driver Behavior within a Signalized Intersection Approach Decision–Dilemma Zone," *Transportation Research Record: Journal of the Transportation Research Board*, vol. 2069, no. 1, pp. 16–25, Jan. 2008, doi: 10.3141/2069-03.

[6] M. Abu Tami, H. I. Ashqar, M. Elhenawy, S. Glaser, and A. Rakotonirainy, "Using Multimodal Large Language Models (MLLMs) for Automated Detection of Traffic Safety-Critical Events," *Vehicles*, vol. 6, no. 3, pp. 1571–1590, 2024.

[7] H. I. Ashqar, A. Jaber, T. I. Alhadidi, and M. Elhenawy, "Advancing Object Detection in Transportation with Multimodal Large Language Models (MLLMs): A Comprehensive Review and Empirical Testing," *arXiv preprint arXiv:2409.18286*, 2024.

[8] M. I. Ashqer *et al.*, "Evaluating a signalized intersection performance using unmanned aerial Data," *Transportation Letters*, vol. 16, no. 5, pp. 452–460, May 2024, doi: 10.1080/19427867.2023.2204249.

[9] S. Lai, Z. Xu, W. Zhang, H. Liu, and H. Xiong, "LLMLight: Large Language Models as Traffic Signal Control Agents," Dec. 2023.

[10] J. Yarlagadda and D. S. Pawar, "Heterogeneity in the Driver Behavior: An Exploratory Study Using Real-Time Driving Data," *J Adv Transp*, vol. 2022, pp. 1–17, Jun. 2022, doi: 10.1155/2022/4509071.

[11] F. Bella and M. Silvestri, "Interaction driver–bicyclist on rural roads: Effects of cross-sections and road geometric elements," *Accid Anal Prev*, vol. 102, pp. 191–201, May 2017, doi: 10.1016/j.aap.2017.03.008.

[12] H. I. Ashqar, T. I. Alhadidi, M. Elhenawy, and N. O. Khanfar, "Leveraging Multimodal Large Language Models (MLLMs) for Enhanced Object Detection and Scene Understanding in Thermal Images for Autonomous Driving Systems," *Automation*, vol. 5, no. 4, pp. 508–526, 2024.

[13] S. Jaradat, R. Nayak, A. Paz, H. I. Ashqar, and M. Elhenawy, "Multitask learning for crash analysis: A Fine-Tuned LLM framework using twitter data," *Smart Cities*, vol. 7, no. 5, pp. 2422–2465, 2024.

[14] S. Masri, H. I. Ashqar, and M. Elhenawy, "Large Language Models (LLMs) as Traffic Control Systems at Urban Intersections: A New Paradigm," *arXiv preprint arXiv:2411.10869*, 2024.

[15] Z. Zhang *et al.*, "Large Language Models for Mobility in Transportation Systems: A Survey on Forecasting Tasks," May 2024.

[16] A. Pang, M. Wang, M.-O. Pun, C. S. Chen, and X. Xiong, "iLLM-TSC: Integration reinforcement learning and large language model for traffic signal control policy improvement," Jul. 2024.

[17] M. Wang, A. Pang, Y. Kan, M.-O. Pun, C. S. Chen, and B. Huang, "LLM-Assisted Light: Leveraging Large Language Model Capabilities for Human-Mimetic Traffic Signal Control in Complex Urban Environments," Mar. 2024.

[18] A. A. Ouallane, A. Bahnasse, A. Bakali, and M. Talea, "Overview of Road Traffic Management Solutions based on IoT and AI," *Procedia Comput Sci*, vol. 198, pp. 518–523, 2022, doi: 10.1016/j.procs.2021.12.279.

[19] H. Almukhalfi, A. Noor, and T. H. Noor, "Traffic management approaches using machine learning and deep learning techniques: A survey," *Eng Appl Artif Intell*, vol. 133, p. 108147, Jul. 2024, doi: 10.1016/j.engappai.2024.108147.

[20] C. Cui, Y. Ma, X. Cao, W. Ye, and Z. Wang, "Receive, Reason, and React: Drive as You Say, With Large Language Models in Autonomous Vehicles," *IEEE Intelligent Transportation Systems Magazine*, vol. 16, no. 4, pp. 81–94, Jul. 2024, doi: 10.1109/MITS.2024.3381793.

[21] L. Chen *et al.*, "Driving with LLMs: Fusing Object-Level Vector Modality for Explainable Autonomous Driving," in *2024 IEEE International Conference on Robotics and Automation (ICRA)*, IEEE, May 2024, pp. 14093–14100. doi: 10.1109/ICRA57147.2024.10611018.

[22] Y. Ren, Y. Chen, S. Liu, B. Wang, H. Yu, and Z. Cui, "TPLLM: A Traffic Prediction Framework Based on Pretrained Large Language Models," Mar. 2024.

[23] D. Zhang, H. Zheng, W. Yue, and X. Wang, "Advancing ITS Applications with LLMs: A Survey on Traffic Management, Transportation Safety, and Autonomous Driving," 2024, pp. 295–309. doi: 10.1007/978-3-031-65668-2_20.

[24] S. Montiel-Marín, C. Gómez-Huélamo, J. de la Peña, M. Antunes, E. López-Guillén, and L. M. Bergasa, "Towards LiDAR and RADAR Fusion for Object Detection and Multi-object Tracking in CARLA Simulator," 2023, pp. 552–563. doi: 10.1007/978-3-031-21062-4_45.

[25] H. Liu, C. Wu, and H. Wang, "Real time object detection using LiDAR and camera fusion for autonomous driving," *Sci Rep*, vol. 13, no. 1, p. 8056, May 2023, doi: 10.1038/s41598-023-35170-z.

[26] C. Gomez-Huelamo, L. M. Bergasa, R. Gutierrez, J. F. Arango, and A. Diaz, "SmartMOT: Exploiting the fusion of HDMaps and Multi-Object Tracking for Real-Time scene understanding in Intelligent Vehicles applications," in *2021 IEEE Intelligent Vehicles Symposium (IV)*, IEEE, Jul. 2021, pp. 710–715. doi: 10.1109/IV48863.2021.9575443.

[27] M. M. Hussien, A. N. Melo, A. L. Ballardini, C. S. Maldonado, R. Izquierdo, and M. Á. Sotelo, "RAG-based Explainable Prediction of Road Users Behaviors for Automated Driving using Knowledge Graphs and Large Language Models," May 2024.

[28] Z. Xu *et al.*, "DriveGPT4: Interpretable End-to-End Autonomous Driving Via Large Language Model," *IEEE Robot Autom Lett*, vol. 9, no. 10, pp. 8186–8193, Oct. 2024, doi: 10.1109/LRA.2024.3440097.

[29] J. Liu, P. Hang, X. Qi, J. Wang, and J. Sun, "MTD-GPT: A Multi-Task Decision-Making GPT Model for Autonomous Driving at Unsignalized Intersections," in *2023 IEEE 26th International Conference on Intelligent Transportation Systems (ITSC)*, IEEE, Sep. 2023, pp. 5154–5161. doi: 10.1109/ITSC57777.2023.10421993.

[30] M. Baeumler and M. Lehmann, "ListDB RepTwo: 3 months (Jun'22-Aug'22) of drone videos and trajectories at a 4-way intersection," Dec. 2023. doi: 10.25532/OPARA-230.